\documentclass[journal]{IEEEtran}

\usepackage[cmex10]{amsmath}
\usepackage{amssymb}

\usepackage{cases}
\usepackage{multirow}
\usepackage{empheq}

\usepackage{cite}
\usepackage{verbatim}

\usepackage{algorithm}
\usepackage{algpseudocode}

\ifCLASSINFOpdf
\usepackage[pdftex]{graphicx}
\else
\usepackage[dvips]{graphicx}
\fi
\usepackage[caption=false,font=footnotesize]{subfig}

\usepackage{color}
\usepackage{multirow}

\usepackage{units}
\usepackage{siunitx}

\newtheorem{theorem}{Theorem}

\newtheorem{corollary}[theorem]{Corollary}

\newtheorem{definition}{Definition}
\newtheorem{remark}{Remark}

\newcommand{\vect}[1]{\mathbf{#1}}

\begin{document}
	
	\sloppy
	
	\title{Boosting Classifiers with Noisy Inference}
	
	
	\author{\IEEEauthorblockN{
			Yongjune Kim, Yuval Cassuto, and Lav R. Varshney}
		
%
%
	
		\thanks{
		Y. Kim was with the Coordinated Science Laboratory, University of Illinois at Urbana-Champaign, Urbana, IL 61801 USA. He is now with Western Digital Research, Milpitas, CA 95035 USA (e-mail: \mbox{yongjune.kim}@wdc.com). Y. Cassuto is with the Viterbi Department of Electrical Engineering, Technion--Israel Institute of Technology, Haifa, Israel (e-mail: \mbox{ycassuto}@ee.technion.ac.il). L. R. Varshney is with the Coordinated Science Laboratory, University of Illinois at Urbana-Champaign, Urbana, IL 61801 USA (e-mail: \mbox{varshney}@illinois.edu).

	  	Lav R. Varshney was supported in part by the National Science Foundation under Grant CCF-1717530. Yuval Cassuto was supported in part by the US-Israel Binational Science Foundation and the Israel Science Foundation.}	
	
	}

	
	\maketitle
	
	\begin{abstract}
		We present a principled framework to address resource allocation for realizing boosting algorithms on substrates with communication or computation noise. Boosting classifiers (e.g., AdaBoost) make a final decision via a weighted vote from the outputs of many base classifiers (weak classifiers). Suppose that the base classifiers' outputs are noisy or communicated over noisy channels; these noisy outputs will degrade the final classification accuracy. We show that this degradation can be effectively reduced by allocating more system resources for more important base classifiers. We formulate resource optimization problems in terms of importance metrics for boosting. Moreover, we show that the optimized noisy boosting classifiers can be more robust than bagging for the noise during inference (test stage). We provide numerical evidence to demonstrate the benefits of our approach. 
	\end{abstract}
	
	\section{Introduction}
	
	
	Boosting methods are machine learning algorithms that construct a set of base (weak) classifiers and then classify a new data point by taking a \emph{weighted} vote of their decisions~\cite{Dietterich2000ensemble}. Boosting can achieve good classification accuracy even if the base classifiers have performance that is only slightly better than random guessing~\cite{Freund1997decision,Bishop2006pattern}. Adaptive boosting (AdaBoost) is the most widely used form of boosting~\cite{Freund1997decision,Freund1996experiments}; it works well for classification problems such as face detection~\cite{Viola2001rapid} and can be extended to regression problems~\cite{Friedman2001greedy}. 
	
	Consider the standard supervised classification problem. For the given training set $S = \{(\vect{x}_1, y_1), \ldots, (\vect{x}_N, y_N)\}$, the objective of learning is to estimate the unknown function $y = f(\vect{x})$ based on the given training set. The input vector is given by $\vect{x}_n = (x_{n,1},\ldots,x_{n,D})$ where $D$ denotes the dimension of the input vectors. The output variables $y_n$ are typically drawn from a discrete set of classes, i.e., $y \in \{1, \ldots, K\}$ where $K$ denotes the number of classes. For a binary classification problem, we assume $y \in \{+1, -1\}$. 
	

		
	The final output of AdaBoost is as follows:
	\begin{equation}\label{eq:weighted_voting}
	f(\vect{x}) = \text{sign} \left(\sum_{t=1}^{T}{\alpha_t f_t(\vect{x})} \right), 
	\end{equation}
	where $\boldsymbol{\alpha} = (\alpha_1, \ldots, \alpha_T)$ denotes the coefficients for base classifiers. AdaBoost assigns larger coefficients to more accurate (or important) base classifiers~\cite{Freund1996experiments,Bishop2006pattern}. Unlike AdaBoost, the coefficients of bagging classifiers are uniform (i.e., $\alpha_t = \frac{1}{T}$ for all $t \in \{1, \ldots, T\}$) and the output of bagging corresponds to majority voting~\cite{Breiman1996bagging}. 
		
	\begin{figure}[t]
		\centering
		\includegraphics[width=0.48\textwidth]{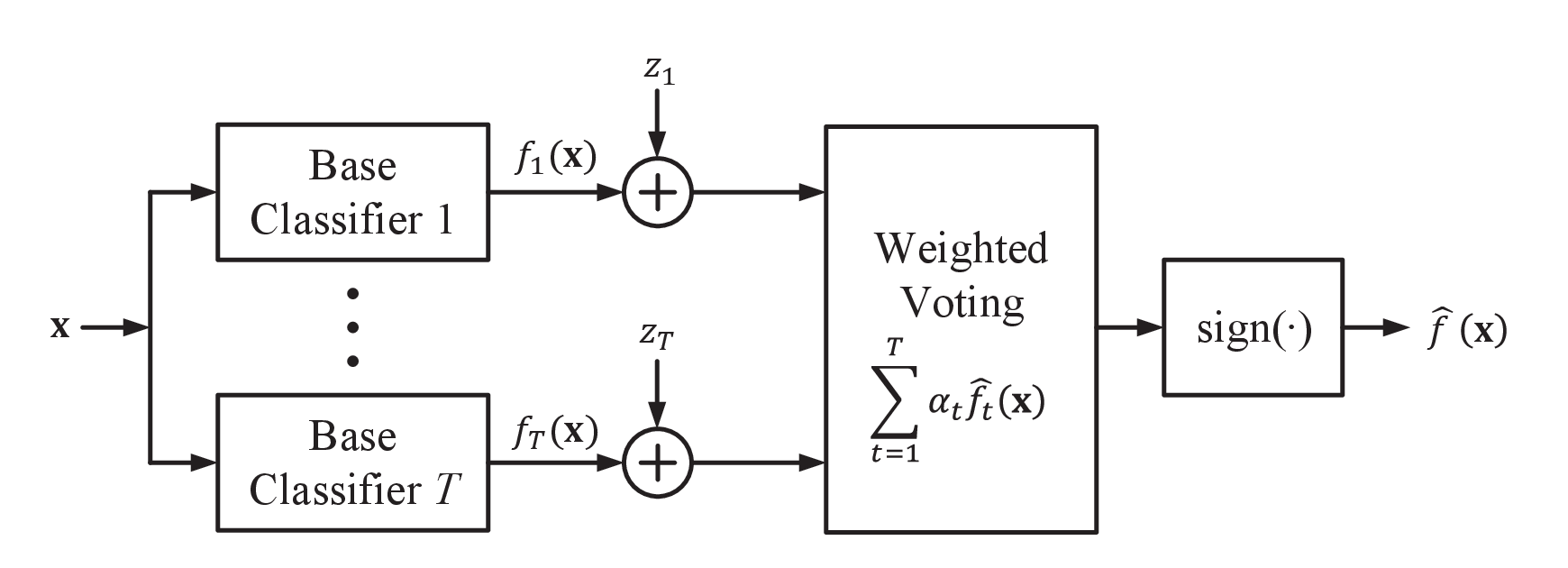}
		\caption{Noisy boosting classifier.}
		\label{fig:noisy_adaboost}
	\end{figure}
				
	Suppose that the outputs of base classifiers are corrupted by random noise as shown in Fig.~\ref{fig:noisy_adaboost}. The noise $\vect{z} = (z_1, \ldots, z_T)$ captures \emph{communication} errors over the channel between the base classifiers and the weighted voter. Alternatively, $\vect{z}$ can originate from noise in the \emph{computation} hardware of base classifiers~\cite{Shanbhag2019shannon}. The corrupted output of the $t$-th base classifier is denoted by $\widehat{f}_t(\vect{x}) \in \{+1, -1\}$. We assume that the weighted vote is implemented in a noiseless manner. 
	
		
	
	We observe that noise in the individual base classifiers affects the overall classification accuracy in a way that strongly depends on the coefficient vector $\boldsymbol{\alpha}$. That is, an erroneous $\widehat{f}_t$ with a large coefficient $\alpha_t$ is more likely to corrupt the final classification output than a base classifier with a smaller coefficient. Following this observation, we develop a principled framework to optimize the classification accuracy by allocating reliability resources to base classifiers according to their importance prescribed in the coefficient vector $\boldsymbol{\alpha}$.       
	
	Ideally, the system resources should be allocated to minimize the classification error probability. However, the classification error probability of boosting depends on the data sets and base classifiers (and their training algorithms); the classification error probability is not simply related to the system resources and does not yield tractable optimization procedures. To circumvent this problem, we minimize \emph{proxies} instead of the classification error probability. First, we define three proxies: 1) Markov proxy, 2) Chernoff proxy, and 3) Gaussian proxy. Next, we formulate optimization problems to minimize these proxies for a given resource budget.  This kind of indirect approach is effective in many engineering problems, e.g.,~\cite{Poor1977applications,Sakr2017analytical,Sakr2017minimum}.
	   
	
	
	In this paper, we assume that the impact of $\vect{z}$ can be controlled by allocating system resources. One example we investigate is that the outputs of base classifiers are corrupted by additive noise over the channels between base classifiers and the weighted voter. Here, the noise level over these channels can be controlled by allocating transmit power. We show the proposed framework can effectively reduce the classification error probability for a given transmit-power budget. 
			
	Our approach provides a general framework to allocate a limited resource for boosting classifiers and can also be applied to settings of noisy computations. For example, the quality of computations on noisy hardware can be changed by controlling supply voltage~\cite{Shanbhag2019shannon}, replicating computations~\cite{Neumann1956probabilistic,Donmez2016cost}, and implementing granular bit precisions~\cite{Sakr2017analytical}. Based on the proposed framework, we can optimize these system resources in a principled manner. For the opposite problem, adversarial attackers can exploit the importance metrics to best degrade the classification accuracy. In this attacking scenario, the attackers should allocate more attacking resource to more important base classifiers. We focus on AdaBoost in this paper, but our approach can be applied to any other weighted ensemble method in machine learning. 	
	
	Our problem of noisy AdaBoost is distinct from AdaBoost in the presence of noisy labels. A well-known model of random classification noise (RCN) assumes that each label $y$ in the training set is flipped independently~\cite{Angluin1988learning,Frenay2014classification}. Several studies have investigated the behavior of AdaBoost under label noise and proposed more robust training algorithms~\cite{Dietterich2000experimental,Domingo2000madaboost,Long2010random}. Note that the \emph{data} noise affects all base classifiers during training; hence, it affects the AdaBoost model (i.e., base classifiers and their coefficients) permanently. Our model assumes that the \emph{system} noise during inference (test stage) affects the decisions of base classifiers independently. We optimize the system resource to mitigate the noise impact without altering the trained AdaBoost models.
	
	It is well known that the classification accuracy of AdaBoost tends to degrade more than that of bagging for the RCN model~\cite{Dietterich2000experimental,Mohri2018foundation}. The reason is that AdaBoost more aggressively fits noisy instances in the training set~\cite{Frenay2014classification,Dietterich2000experimental}. Contrarily, we show that AdaBoost can be more robust than bagging in our problem setting where noise flips the base classifiers' outputs during inference (test). This is because the accuracy improvement by the proposed optimization is more effective as the coefficient variability increases, or formally, as the geometric mean of the coefficients decreases. 	
	
	Our noise model is different from the model in \emph{error-aware} inference~\cite{Wang2015eacb,Wang2015overcoming}. Error-aware inference is a retraining approach to overcome computational errors due to hardware non-idealities. The basic idea is to retrain (i.e., update the trained models) by taking into account noisy computations~\cite{Shanbhag2019shannon}. However, this retraining approach only works for permanent noise (e.g., stuck-at faults) since random (transient) noise cannot be trained. In this paper, we focus on random errors. 
	
	
	If we regard base classifiers as distributed sensors, then the AdaBoost's weighted voting is similar to the Bayesian fusion rule of \emph{distributed detection} with multiple sensors~\cite{Chair1986optimal,Viswanathan1997distributed}. In distributed detection, each local sensor's optimal decision rule is the likelihood ratio test (LRT) for conditionally independent sensor observations~\cite{Viswanathan1997distributed}, and usually the observation of each local sensor is a scalar value. On the other hand, AdaBoost can use any learning algorithms that train their models based on high-dimensional data sets. For example, decision trees~\cite{Dietterich2000experimental,Viola2001rapid}, support vector machines~\cite{Li2008adaboost}, and neural networks~\cite{Schwenk1997training,Schwenk2000boosting} have been investigated for base classifiers of AdaBoost. 
	
	Our noisy inference problem is distinct from \emph{channel-aware} distributed detection with multiple sensors~\cite{Chen2005optimality,Chen2006channel}. The distinction lies in that our approach controls the channel distributions whereas channel-aware distributed detection attempts to optimize the thresholds of each sensor's LRT for fixed channel distributions. In addition, the approach of channel-aware distributed detection does not work for AdaBoost because 1) the distribution of $\vect{x}$ is not known in learning problems, 2) base classifiers are not simple LRTs, and 3) AdaBoost does not require the conditional independence assumption among base classifiers unlike distributed detection with multiple sensors.

		
	The rest of this paper is organized as follows. Section~\ref{sec:adaboost} explains the noisy AdaBoost model. Section~\ref{sec:proxy} develops three metrics for the importance of base classifiers originating from three optimization problems. Section~\ref{sec:resource} formulates and solves resource-allocation problems based on these importance metrics. Section~\ref{sec:numerical} provides numerical results and Section~\ref{sec:conclusion} concludes.

	\section{Noisy AdaBoost Model}\label{sec:adaboost}
		
	\subsection{AdaBoost}
	
	AdaBoost trains the base classifiers in sequence to minimize an exponential error function~\cite{Bishop2006pattern,Freund1996experiments}. Each base classifier is trained using a weighted form of the training set in which the data weights $\vect{w} =(w_1, \ldots, w_N)$ depend on the performance of previous base classifiers. In particular, data points that are misclassified by one of the base classifiers are given greater weight when used to train the next base classifier. Once all base classifiers have been trained, their outputs are combined through weighted voting~\cite{Bishop2006pattern}. 
	
	Note that the data weights $\vect{w}= (w_1, \ldots, w_N)$ are distinct from the classifier coefficients $\boldsymbol{\alpha} = (\alpha_1, \ldots, \alpha_T)$. AdaBoost determines both values during training. Once training is done, only the coefficients $\boldsymbol{\alpha}$ are used to classify new data points. The training of AdaBoost is given by Algorithm~\ref{algo:adaboost}.  
	
	\begin{algorithm}
		\caption{Training of AdaBoost for binary classification~\cite{Bishop2006pattern}} \label{algo:adaboost}
		\begin{algorithmic}[1]
			\State Initialize the data weights $\vect{w}$ by setting $w_n^{(1)} = \frac{1}{N}$. 
			\For{$t = 1:T$}
				\State Fit a base classifier $f_t (\vect{x})$ to the training set by minimizing the weighted error function
				\begin{equation} \label{eq:weighted_cost}
					J_t = \sum_{n=1}^{N}{w_n^{(t)} \mathcal{I}(f_t (\vect{x}_n) \ne y_n)}
				\end{equation}
				where $\mathcal{I}(f_t (\vect{x}_n) \ne y_n)$ denotes the indicator function and equals $1$ if $f_t (\vect{x}_n) \ne y_n$ and $0$ otherwise. 
				\State Evaluate
				\begin{equation}\label{eq:epsilon}
				\varepsilon_t = \frac{\sum_{n=1}^{N}{w_n^{(t)}  \mathcal{I}(f_t(\vect{x}_n) \ne y_n)}}{\sum_{n=1}^{N}{w_n^{(t)}}}. 
				\end{equation}
				\State Compute the classifier coefficients
				\begin{equation}
				\alpha_t = \log{\frac{1 - \varepsilon_t}{\varepsilon_t}}. 
				\end{equation}
				\State Update the data weights
				\begin{equation}
				w_n^{(t+1)} = w_n^{(t)}\exp{\{\alpha_t \mathcal{I}(f_t(\vect{x}_n) \ne y_n)\}}. 
				\end{equation}
			\EndFor
			\State \textbf{return} the trained base classifiers $\{f_t(\cdot)\}$ and the corresponding coefficients $\boldsymbol{\alpha}$ for $t \in \{1,\ldots, T\}$. 
		\end{algorithmic}
	\end{algorithm}
	
	The final model of AdaBoost is given by~\eqref{eq:weighted_voting} where the base classifiers and the coefficients are decided by Algorithm~\ref{algo:adaboost}. The classification error probability of the trained model $f(\cdot)$ is given by
	\begin{equation}
	P_{e,f} = \Pr(f(\vect{x}) \ne y), 
	\end{equation}
	where $y$ is the true label corresponding to $\vect{x}$. 	
	
	\begin{remark}[Positive Coefficients]\label{rem:positive_a} If a base classifier is better than random guessing, then $\alpha_t > 0$ for any $t \in \{1, \ldots, T\}$~\cite{Freund1997decision}. 
	\end{remark}
	
	\begin{remark}[Normalized Coefficients]\label{rem:normalization} We normalize the coefficients such that $\sum_{t=1}^{T}{\alpha_t}=1$. Note that normalization does not affect the classification output in~\eqref{eq:weighted_voting}.
	\end{remark}

	\begin{remark}[Distinction from Distributed Detection Problem]
		We note that $\varepsilon_t$ in \eqref{eq:epsilon} depends on the data weights $\vect{w}$ unlike the distributed detection problem in \cite{Viswanathan1997distributed}. 
	\end{remark}

	
	\subsection{Noisy AdaBoost}
	
	Suppose that the base classifiers' outputs may be flipped due to the noise $z_t$, i.e., $f_t(\vect{x}) \ne \widehat{f}_t(\vect{x})$ where 
	\begin{equation}\label{eq:noisy_base}
	\widehat{f}_t(\vect{x}) = \text{sign} \left(f_t(\vect{x}) + z_t \right). 
	\end{equation} 	
	The mismatch event of the $t$-th base classifier is denoted by
	\begin{equation}
	\delta_{t} = \mathcal{I}(f_t(\vect{x})\ne\widehat{f}_t(\vect{x})). 
	\end{equation}
	Then, we can define the base classifiers' mismatch probabilities as $\vect{p} = \left(p_1, \ldots, p_T\right)$ where
	\begin{equation} \label{eq:mismatch_base}
	p_t \triangleq \Pr(f_t(\vect{x}) \ne \widehat{f}_t(\vect{x})) = \mathbb{E}[\delta_t]. 
	\end{equation}
	In the sequel, the expectation over the distribution of $\vect{x}$ will be replaced by the empirical mean over the given data set. 
	
	The final output of noisy AdaBoost is given by
	\begin{equation}\label{eq:noisy_weighted_voting}
	\widehat{f}(\vect{x}) = \text{sign} \left(\sum_{t=1}^{T}{\alpha_t \widehat{f}_t(\vect{x})} \right). 
	\end{equation}
	Then, the final mismatch probability (i.e., mismatch probability of the final output) is given by 
	\begin{equation} \label{eq:mismatch}
	P_m \triangleq \Pr(f(\vect{x}) \ne \widehat{f}(\vect{x})), 
	\end{equation}
	which captures the negative impact of $\vect{z}$ on the final classification accuracy. We can expect that the final mismatch probability $P_m$ depends on the base classifiers' mismatch probabilities $\vect{p}$. 
	
	The classification error probability of the noisy AdaBoost is upper bounded by 
	\begin{equation}\label{eq:mismatch_bound}
	P_e = P_{e,\widehat{f}} \le P_{e,f} + P_{m},
	\end{equation}
	where $P_{e,f}$ denotes the classification error probability by the noise-free AdaBoost. In~\cite{Sakr2017analytical,Sakr2017minimum}, the mismatch probability characterizes the impact of quantization noise due to limited bit precision. Note that $P_{e,f}$ solely depends on the AdaBoost algorithm and the dataset, i.e., $P_{e,f}$ is independent of $\vect{z}$. Hence, we focus on $P_m$ to reduce the negative impact of the noise $\vect{z}$. 
		
	\section{Importance Metrics of Base Classifiers}\label{sec:proxy}
	
	We define three proxies to the mismatch probability: 1) Markov proxy, 2) Chernoff proxy, and 3) Gaussian proxy. These proxies induce different importance metrics of base classifiers. We	provide theoretical justification for the proxies and the corresponding metrics. 
	
	\subsection{Markov Proxy}
	
	Here, we define the \emph{Markov proxy}, which comes from Markov's inequality. 
		
	\begin{definition}[Markov Proxy] \label{def:proxy_markov}
		The Markov proxy $\widehat{p}_M$ of the noisy AdaBoost is given by
		\begin{equation} \label{eq:proxy_markov}
			\widehat{p}_M =  \sum_{t=1}^{T}{\alpha_t p_{t}}, 
		\end{equation}
		which is the nonnegative weighted sum of $p_t$.  
	\end{definition}

	We derive an upper bound on the mismatch probability $P_m$ based on Markov's inequality and show that this upper bound can be lowered by minimizing the Markov proxy $\widehat{p}_M$. 


	\begin{theorem}[Upper Bound by Markov's Inequality]\label{thm:markov_bound} The mismatch probability of $\vect{x}_n$ is upper bounded as follows:
		\begin{equation} \label{eq:markov_bound}
		P_m(\vect{x}_n) \le \frac{2 \widehat{p}_M}{\gamma_n},
		\end{equation}
		where 
		\begin{equation} \label{eq:gamma}
		\gamma_n = \left| \sum_{t=1}^{T}{\alpha_t f_t (\vect{x}_n)} \right|,
		\end{equation}
		which represents the decision margin of $\vect{x}_n$. 
		Then, an upper bound on the mismatch probability $P_m$ is given by
		\begin{equation}
		P_m \le \left( \frac{2}{N} \sum_{n=1}^{N}{\frac{1}{\gamma_n}} \right) \cdot \widehat{p}_M. 
		\end{equation}
	\end{theorem}
	\begin{IEEEproof}
	The proof is given in Appendix~\ref{pf:markov_bound}. 
	\end{IEEEproof}		


	Higher decision margin $\gamma_n$ and/or lower Markov proxy $\widehat{p}_M$ reduce the upper bound on the mismatch probability. The margin $\gamma_n$ depends only on the input vector $\vect{x}_n$ and the trained AdaBoost classifier model. In contrast, $\widehat{p}_M$ depends on $\vect{z}$, whose distribution we can control by resource allocation; hence minimizing $\widehat{p}_M$ is pursued in Section \ref{sec:resource}. 
	
	\begin{remark}
	For a given dataset and trained AdaBoost classifier model, the upper bound~\eqref{eq:markov_bound} depends only on the Markov proxy $\widehat{p}_M$. Hence, our objective in Section~\ref{sec:resource} is to minimize $\widehat{p}_M$ by controlling $\vect{p} = (p_1, \ldots, p_T)$. 
	\end{remark}


%

	\subsection{Chernoff Proxy}
	
	In this subsection, we define the \emph{Chernoff proxy} via the Chernoff bound. 
	
	\begin{definition}[Chernoff Proxy]\label{def:proxy_chernoff} The Chernoff proxy $\widehat{p}_C$ of the mismatch probability is given by
		\begin{equation} \label{eq:proxy_chernoff}
		\widehat{p}_C(s) = \sum_{t=1}^{T}{ \left(e^{s \alpha_t} - 1\right)p_t },
		\end{equation}
		where $s > 0$ is a parameter. 
	\end{definition}

	\begin{remark}\label{rem:positive_b}
		Since $\alpha_t > 0$ and $s>0$, $e^{s \alpha_t} - 1 > 0$.
	\end{remark}
	
	We derive an upper bound on the mismatch probability $P_m$ from the Chernoff bound and show that this upper bound can be reduced by minimizing the Chernoff proxy $\widehat{p}_C$. 
	
	\begin{theorem}[Upper Bound by Chernoff Bound]\label{thm:chernoff_bound}
		The mismatch probability is upper bounded as follows:
		\begin{equation} \label{eq:chernoff_bound}
		P_m \le \mathbb{E}\left[ \exp\left(-s \cdot \frac{\gamma_n}{2}\right) \right] \cdot \exp\left( \widehat{p}_C(s) \right),
		\end{equation}
		for any $s > 0$. 
	Note that $\mathbb{E}\left[ \exp\left(-s \cdot \frac{\gamma_n}{2}\right) \right]$ can be calculated by
	\begin{equation}
	\mathbb{E}\left[ \exp\left(-s \cdot \frac{\gamma_n}{2}\right) \right] = \frac{1}{N} \sum_{n=1}^{N}{\exp\left(-s \cdot \frac{\gamma_n}{2}\right)}. 
	\end{equation}
	\end{theorem}
	\begin{IEEEproof}
		The proof is given in Appendix~\ref{pf:chernoff_bound}. 
	\end{IEEEproof}

	This upper bound can be tightened by minimizing the Chernoff proxy $\widehat{p}_C$. Also, similarly to the Markov proxy, a higher decision margin $\gamma_n$ decreases the upper bound. 
	
	In addition, $s$ should be carefully chosen because of a trade-off relation between $\mathbb{E}\left[ \exp\left(-s \cdot \frac{\gamma_n}{2}\right) \right]$ and $\widehat{p}_C$. A smaller $s$ decreases $\widehat{p}_C$ while increasing $\mathbb{E}\left[ \exp\left(-s \cdot \frac{\gamma_n}{2}\right) \right]$. Since the optimal $s$ and $\vect{p}$ are interdependent, we propose an iterative algorithm to jointly find the optimal $s$ and $\vect{p}$ (see Algorithm~\ref{algo:chernoff_bound} in Section~\ref{sec:resource}).        
		
	\subsection{Gaussian Approximation} \label{sec:proxy_gaussian}
	
	As in Definition~\ref{def:proxy_markov} and Definition~\ref{def:proxy_chernoff}, we define a third proxy metric that is a nonnegative weighted sum of $p_t$. 

	\begin{definition}[Gaussian Proxy]\label{def:proxy_gauss} The Gaussian proxy $\widehat{p}_G$ of the mismatch probability is given by 
		\begin{equation} \label{eq:proxy_gauss}
		\widehat{p}_G =  \sum_{t=1}^{T}{\alpha_t^2 p_{t}}. 
		\end{equation}
	\end{definition}

	Suppose that $\widehat{f}(\vect{x}_n) = \text{sign}(\widehat{g}(\vect{x}_n))$ where $\widehat{g}(\vect{x}_n)$ is given by
	\begin{align}
	\widehat{g}(\vect{x}_n) &= \sum_{t=1}^{T}{\alpha_t \widehat{f}_t(\vect{x}_n)} \\
	&= \sum_{t\in \mathcal{T}^+_n}{\alpha_t \widehat{f}_t(\vect{x}_n)} + \sum_{t\in \mathcal{T}^-_n}{\alpha_t \widehat{f}_t(\vect{x}_n)} \\
	& = \sum_{t\in \mathcal{T}^+_n}{\alpha_t (1 - 2 \delta_{t,n})} + \sum_{t\in \mathcal{T}^-_n}{\alpha_t (-1 + 2 \delta_{t,n})} \label{eq:noisy_f} \\
	& = \pm \gamma_n + v_n \label{eq:signal_noise}
	\end{align} 
	where $\mathcal{T}^+_n = \{t \mid f_t(\vect{x}_n) = 1 \}$ and $\mathcal{T}^-_n = \{t \mid f_t(\vect{x}_n) = -1 \}$ $\mathcal{T}^-_n$, respectively. Note that $\delta_{t,n} = \mathcal{I}(f_t(\vect{x}_n)\ne\widehat{f}_t(\vect{x}_n))$.
		
	We set $\pm \gamma_n$ and $v_n$ in \eqref{eq:signal_noise} as \emph{signal} term and the \emph{noise} term, respectively. The noise term $v_n$ is given by
	\begin{equation}
	v_n = - 2 \left(\sum_{t\in \mathcal{T}^+_n}{\alpha_t \delta_{t,n}} - \sum_{t\in \mathcal{T}^-_n}{\alpha_t \delta_{t,n}} \right). 
	\end{equation} 
	
	\begin{theorem} \label{thm:gaussian}
		The noise term $v_n$ for $n \in \{1,\ldots,N\}$ can be modeled as a Gaussian distribution, i.e., $v_n \sim \mathcal{N}(\mu_v, \sigma_v^2)$ by the central limit theorem. Then, 
	\begin{align}
	\mu_v & = -2 \left( \sum_{t\in \mathcal{T}^+_n}{\alpha_t p_t} - \sum_{t\in \mathcal{T}^-_n}{\alpha_t p_t} \right), \label{eq:v_mean} \\
	\sigma_v^2 &= 4 \sum_{t=1}^{T}{\alpha_t^2 p_t (1 - p_t)}. \label{eq:v_var}
	\end{align}
	\end{theorem}
	\begin{IEEEproof}
	The proof is given in Appendix~\ref{pf:gaussian}. 	
	\end{IEEEproof}

	Observe that the variance in \eqref{eq:v_var} is \emph{data independent}, and thus its minimization by resource allocation is an effective way to reduce the classification error probability. 

	\begin{remark}
		 If $p_t \ll 1$ (i.e., $p_t^2 \ll p_t$), then $\sigma_v^2 \approx 4 \widehat{p}_G$. The advantage of Gaussian proxy is the convexity of the right-hand side of \eqref{eq:proxy_gauss} (see Section~\ref{sec:resource}). 
	\end{remark}

	Based on the Gaussian approximation, we can derive an estimate of the mismatch probability $P_m$. 
	\begin{corollary}\label{thm:est_gauss} An estimate of the mismatch probability is given by
		\begin{equation} \label{eq:est_gauss}
		P_m (\vect{x}_n) \approx Q\left( \frac{ \gamma_n - \mu_v}{2\sqrt{\widehat{p}_G}} \right),
		\end{equation}
		where $Q(x) = \frac{1}{\sqrt{2\pi}} \int_{x}^{\infty}{\exp\left(-\frac{u^2}{2}\right) du}$. 
	\end{corollary}

	\begin{table}[!t]
	\renewcommand{\arraystretch}{1.4}
	\caption{Comparison of Importance Metrics of Base Classifiers}
	\vspace{-2mm}
	\label{tab:beta}
	\centering
	\begin{tabular}{|c|c|c|}	\hline
		Proxy      & Importance metric $\beta$  & Remarks \\ \hline \hline
		Markov     & $\alpha$  &  Definition~\ref{def:proxy_markov} \\ \hline
		Chernoff   & $e^{s \alpha} - 1$  & Definition~\ref{def:proxy_chernoff} \\ \hline
		Gaussian   & $\alpha^2$ & Definition~\ref{def:proxy_gauss} \\ \hline			
	\end{tabular}
	\end{table} 

	The minimum $\mu_v$ and the minimum $\widehat{p}_G$ are desired to reduce the estimate of mismatch probability. However, $\mu_v$ depends on the input vector $\vect{x}_n$, hence, we cannot easily minimize $\mu_v $. In contrast, the Gaussian proxy depends only on the trained $\boldsymbol{\alpha}$ and the base classifiers' mismatch probability $\vect{p}$. Hence, we can minimize the Gaussian proxy to reduce the mismatch probability. 
	
	Note that each of the three proxies can be described by
	\begin{equation}
	\sum_{t=1}^{T}{\beta_t p_t}, 
	\end{equation}
	where $\beta_t$ denotes the \emph{importance metric} of the $t$-th base classifier. Table~\ref{tab:beta} lists the importance metrics for the three proxies. 
	
\begin{figure}[!t]
		\centering
		\includegraphics[width=0.4\textwidth]{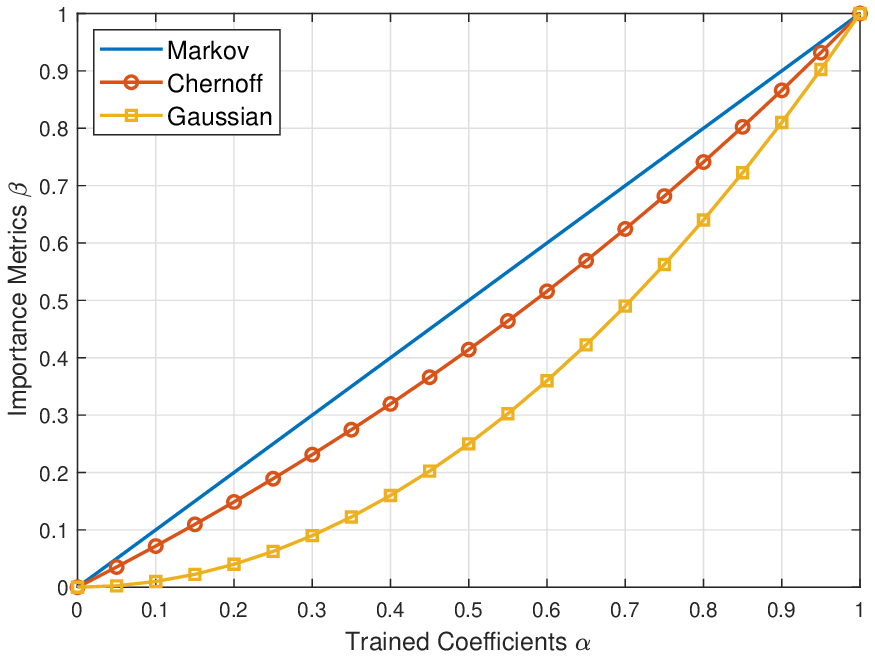}
		\caption{Comparison of importance metrics $\beta$ in Table~\ref{tab:beta} (with $s = \log{2}$ for the Chernoff proxy).}
		\label{fig:beta}
	\end{figure}

	Fig.~\ref{fig:beta} plots the dependence of each importance metric on $\alpha$. It illustrates how resource allocation based on $\beta$ would give preference to a larger $\alpha$. Note that the importance metrics of Markov proxy and Gaussian proxy correspond to $\ell_1$-norm penalty and $\ell_2$-norm penalty, respectively. Hence, the Gaussian proxy puts very small weight on less important base classifiers, but strong weight on more important base classifiers.

	
	\begin{remark} \label{rem:positive_beta}
	The proposed importance metrics are positive (i.e., $\beta_t > 0$) because of $\alpha_t > 0$ (Remark~\ref{rem:positive_a}) and $e^{s \alpha_t} - 1 > 0$ (Remark~\ref{rem:positive_b}).
	\end{remark}

	\section{Resource Allocation for Noisy AdaBoost}\label{sec:resource}
	
	\subsection{Formulation of Optimization Problems}	
	
	We investigate optimization approaches to determine the optimal $\vect{p} = (p_1, \ldots, p_T)$ for a given resource constraint. By optimizing the proposed proxies, we attempt to reduce the mismatch probability, i.e., reduce the noise impact on classification accuracy. 
	
	An important assumption is that the mismatch probabilities of base classifiers can be controlled by allocating the resources. Suppose that the mismatch probability of the $t$-th base classifier $p_t$ can be described by resource $r_t$, i.e., $p_t = p(r_t)$. Then, we can formulate the following optimization problem for a given resource budget $\mathcal{C}$: 
	\begin{equation}
	\begin{aligned} \label{eq:opt_proxy}
	& \underset{\vect{r}}{\text{minimize}}
	& & \sum_{t=1}^{T}{\beta_t p(r_t)} \\
	&{\text{subject~to}} & & \sum_{t=1}^{T}{c(r_t)} \le \mathcal{C} 
	\end{aligned}
	\end{equation}
	where the objective function depends on the importance metric $\boldsymbol{\beta} = (\beta_1, \ldots, \beta_T)$. Also, $c(r_t)$ denotes the cost of the allocated resource to the $t$-th base classifier.
	
	\begin{remark}
		If $p(r_t)$ and $c(r_t)$ are \emph{convex}, then the optimization problem \eqref{eq:opt_proxy} is also \emph{convex} because $\beta_t$ is positive for all $t$ in any of the three proxies (Remark \ref{rem:positive_beta}). In such cases, for the Markov proxy and the Gaussian proxy, we can obtain the optimal resource allocation by solving \eqref{eq:opt_proxy} directly using convex programming. 
	\end{remark}

	For the Chernoff proxy \eqref{eq:proxy_chernoff}, due to the free parameter $s$, we propose an iterative algorithm to jointly find the optimal $s$ and $\vect{p}$ (see Algorithm~\ref{algo:chernoff_bound}).        
	
	\begin{algorithm}
		\caption{Iterative algorithm to minimize the upper bound in Theorem~\ref{thm:chernoff_bound}} \label{algo:chernoff_bound}
		\begin{algorithmic}[1]
			\State Choose an arbitrary starting point $s^{(0)}$ in $(0, \infty)$ and set $i=0$. 
			\Repeat
			\State Step 1. Solve the following optimization problem:
			\begin{equation} \label{eq:algo_p}
			\vect{p}^{(i+1)} = \arg\min_{\vect{p}} \widehat{p}_C(s^{(i)}). 
			\end{equation}
			\State Step 2. Find $s^{(i+1)}$ satisfying
			\begin{equation} \label{eq:algo_s}
			\sum_{n=1}^{N}{\left\{ \left(\sum_{t=1}^{T}{p_t^{(i+1)}\alpha_t e^{s \alpha_t}} - \frac{\gamma_n}{2}\right) \cdot e^{-s \cdot \frac{\gamma_n}{2}}\right\}} = 0.
			\end{equation}
			\State Step 3. $s^{(i+1)} = \max\left\{s^{(i+1)}, \epsilon\right\}$. 
			\State Step 4. $i \gets i+1$. 
			\Until{stopping criterion is satisfied.}  
		\end{algorithmic}
	\end{algorithm}	

	Algorithm~\ref{algo:chernoff_bound} attempts to minimize the upper bound of \eqref{eq:chernoff_bound} by alternating between optimizations of \eqref{eq:algo_p} and \eqref{eq:algo_s}. Step 1 finds $\vect{p}^{(i+1)}$ minimizing $\widehat{p}_C(s^{(i)})$ for a given $s^{(i)}$, which is a convex problem for any $s^{(i)} > 0$. It is because $\vect{p}$ affects only $\widehat{p}_C(s)$ among the upper bound of \eqref{eq:chernoff_bound}, which corresponds to \eqref{eq:opt_proxy}. Step 2 finds in closed form $s^{(i+1)}$ minimizing the upper bound of \eqref{eq:chernoff_bound} for a given $\vect{p}^{(i+1)}$. Step 3 introduces a small positive $\epsilon > 0 $ to satisfy the condition of $s > 0$.

	We show that the upper bound is a \emph{convex} function of $s$ and the solution of \eqref{eq:algo_s} is optimal.

	\begin{theorem}\label{thm:optimal_s}
		For given $\vect{p} = (p_1, \ldots, p_T)$ and ${\boldsymbol \gamma}=(\gamma_1, \ldots, \gamma_N)$, the upper bound of \eqref{eq:chernoff_bound} is a function of $s$ as follows: 
		\begin{equation} \label{eq:h}
		h(s) = \frac{1}{N} \sum_{n=1}^{N}{e^{-s \cdot \frac{\gamma_n}{2}}} \cdot \exp\left(\sum_{t=1}^{T}{\left( e^{s \alpha_t} - 1\right)p_t}\right).   
		\end{equation}  
		This upper bound function $h(s)$ is \emph{convex}; hence $s$ satisfying $h'(s)=0$ (i.e., \eqref{eq:algo_s}) is optimal. 
	\end{theorem}
	\begin{IEEEproof}
		The proof is given in Appendix~\ref{pf:optimal_s}.
	\end{IEEEproof}
	
	\begin{corollary}
		If $\gamma_n = \gamma_0$ for all $n\in\{1,\ldots, N\}$ and $\alpha_t = \frac{1}{T}$ for all $t\in\{1,
		\ldots,T\}$, then the optimal $s$ is given by
		\begin{equation}
		s^* = \max \left\{T \cdot \log{\frac{\gamma_0}{2 \cdot \widetilde{p}}}, \epsilon \right\}
		\end{equation}
		where $\widetilde{p} = \tfrac{\sum_{t=1}^{T}{p_t}}{T}$. 
		\begin{IEEEproof}
			If $\gamma_n = \gamma_0$ for any $n$ and $\alpha_t = \frac{1}{T}$ for any $t$, then \eqref{eq:h} is given by
			\begin{equation}
			h(s) = \exp\left( -\frac{\gamma_0}{2} s + \sum_{t=1}^{T}{(e^{\frac{s}{T}} -1) p_t}  \right). 			
			\end{equation}
			Then, the minimization of $h(s)$ is equivalent to minimizing $q(s) = -\frac{\gamma_0}{2} s + \sum_{t=1}^{T}{(e^{\frac{s}{T}} -1) p_t}$. It is clear that $s^* = T \cdot \log{\frac{\gamma_0}{2 \cdot \widetilde{p}}}$ satisfies $q'(s) = 0$. If $s^* \le 0$, then it is replaced by $\epsilon$. 			
		\end{IEEEproof}		
	\end{corollary}
	Note that a larger noise margin $\gamma_0$ increases the optimal $s$ whereas a larger $\widetilde{p}$ reduces the optimal $s$.

	
	\subsection{Example: Communication Power Allocation}
	
	Suppose that the $f_t(\vect{x}) \in \{+1, -1\}$ is transmitted by a symbol from $\{r_t , -r_t\}$, which is corrupted by the noise $z_t$ as shown in Fig.~\ref{fig:noisy_adaboost}. In many applications, the additive noise can be modeled as Gaussian distribution, i.e., $z_t \sim \mathcal{N}(0, \sigma_t^2)$. Then, 
	\begin{equation}
	p(r_t) = Q\left( \sqrt{\mathsf{SNR}_t} \right) = Q\left( \frac{r_t}{\sigma_t} \right),
	\end{equation}
	where the signal-to-noise ratio (SNR) is $\mathsf{SNR}_t = \frac{r_t^2}{\sigma_t^2}$. Hence, $p_t$ can be controlled by allocating transmit power $c(r_t) = r_t^2$. Then, the optimization problem \eqref{eq:opt_proxy} will be  
	\begin{equation}
	\begin{aligned} \label{eq:opt_communication}
	& \underset{\vect{r}}{\text{minimize}}
	& & \sum_{t=1}^{T}{\beta_t Q\left(\frac{r_t}{\sigma_t}\right)} \\
	&{\text{subject~to}} & & \sum_{t=1}^{T}{r_t^2} \le \mathcal{C} \\
	&                    & & r_t \ge 0 \quad t=1,\ldots,T
	\end{aligned}
	\end{equation}
	where $\mathcal{C}$ represents the total power budget. 
	
	\begin{remark}
		The power allocation problem \eqref{eq:opt_communication} is a convex optimization problem since $p(r_t) = Q(r_t)$ is convex for $r_t \ge 0$. Note that $\frac{d^2Q(x)}{dx^2} = \frac{x}{\sqrt{2\pi}}\exp\left(- \frac{x^2}{2}\right)\ge 0$. 
	\end{remark}

	\begin{theorem}\label{thm:opt_communication_sol}The optimal solution $\vect{r}^*$ of \eqref{eq:opt_communication} is
		\begin{equation} \label{eq:opt_communicatino_sol}
		\mathsf{SNR}_t^* =  \left(\frac{r_t^*}{\sigma_t}\right)^2 = W\left(\frac{\beta_t^2}{8 \pi \sigma_t^4 \nu^2 }\right),
		\end{equation}
		where $\nu$ is a positive dual variable of the Karush-Kuhn-Tucker (KKT) conditions. Also, $W(\cdot)$ denotes the \emph{Lambert W function} (i.e., the inverse function of $f(x) = xe^x$)~\cite{Corless1996lambert}. 
	\end{theorem}
	\begin{IEEEproof}
		The proof is given in Appendix~\ref{pf:opt_communication_sol}. 
	\end{IEEEproof}

	Since $W(x)$ is an increasing function for $x \ge 0$, \eqref{eq:opt_communicatino_sol} shows that as desired we allocate higher SNR for higher $\beta_t$. For a classifier with $\beta_t \simeq 0$, the corresponding SNR is $\mathsf{SNR}_t^* \simeq 0$ because $W(0) = 0$. 

	\begin{corollary}\label{thm:nonuniform}
		If $\sigma_t = \sigma$ for all $t\in \{1,\ldots,T\}$, then the \emph{optimized} proxy of \eqref{eq:opt_communication} can be approximated as:
		\begin{equation}
		\sum_{t=1}^{T}{\beta_t Q\left(\frac{r_t^*}{\sigma_t}\right)} \approx \frac{T}{2}\exp\left(-\frac{\mathcal{C}}{2T\sigma^2}\right) \left( \prod_{t=1}^{T}{\beta_t} \right)^{\frac{1}{T}}
		\end{equation}
		where $\left( \prod_{t=1}^{T}{\beta_t} \right)^{\frac{1}{T}}$ is the geometric mean of $\boldsymbol{\beta}$. 
	\end{corollary}
	\begin{IEEEproof}
		The proof is given in Appendix~\ref{pf:nonuniform}. 
	\end{IEEEproof}

	We observe that a smaller geometric mean of $\boldsymbol{\beta}$ implies a lower proxy value. Note that higher SNR $\frac{\mathcal{C}}{\sigma^2}$ decreases the proxy value. 

	\begin{remark}
	The geometric mean of $\boldsymbol{\beta}$ is maximized for the uniform $\boldsymbol{\alpha} = \left(\frac{1}{T}, \ldots, \frac{1}{T}\right)$. Thus the non-uniform coefficients of AdaBoost's classifiers contribute to lower classification error probability. This suggests an advantage of AdaBoost over bagging that assigns the same coefficients to all classifiers (i.e., $\boldsymbol{\alpha} = \left(\frac{1}{T}, \ldots, \frac{1}{T}\right)$). This is a noteworthy fact because AdaBoost is known to be less robust than bagging in the problem of noisy data labels~\cite{Frenay2014classification,Dietterich2000experimental}.  
	\end{remark}

	\section{Numerical Results}\label{sec:numerical}
	
	\begin{figure}[t]
	\centering
	\subfloat[]{\includegraphics[width=0.4\textwidth]{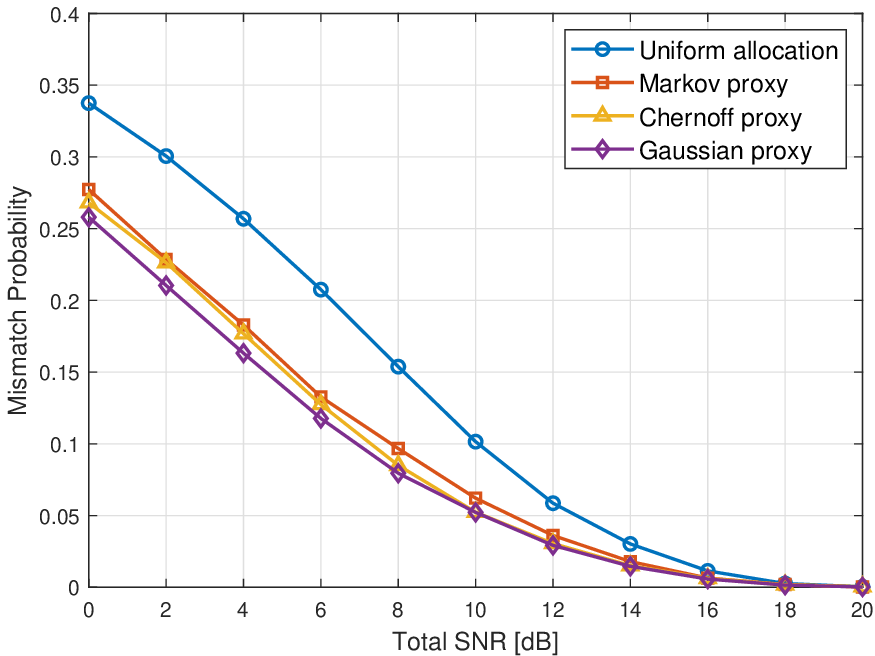}
		\label{fig:T10_pm}}
	\hfil
	\subfloat[]{\includegraphics[width=0.4\textwidth]{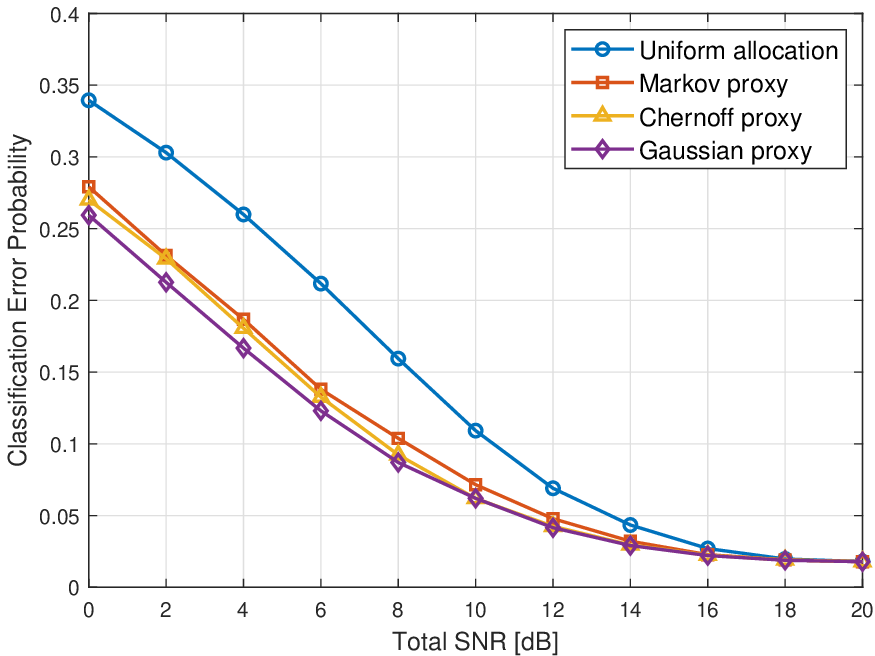}
		\vspace{-4mm}
		\label{fig:T10_pe}}
	\caption{Evaluation of optimized SNRs ($T = 10$): (a) Mismatch probability and (b) classification error probability.}
	\label{fig:T10}
	\end{figure}

	\begin{figure}[t]
	\centering
	\subfloat[]{\includegraphics[width=0.4\textwidth]{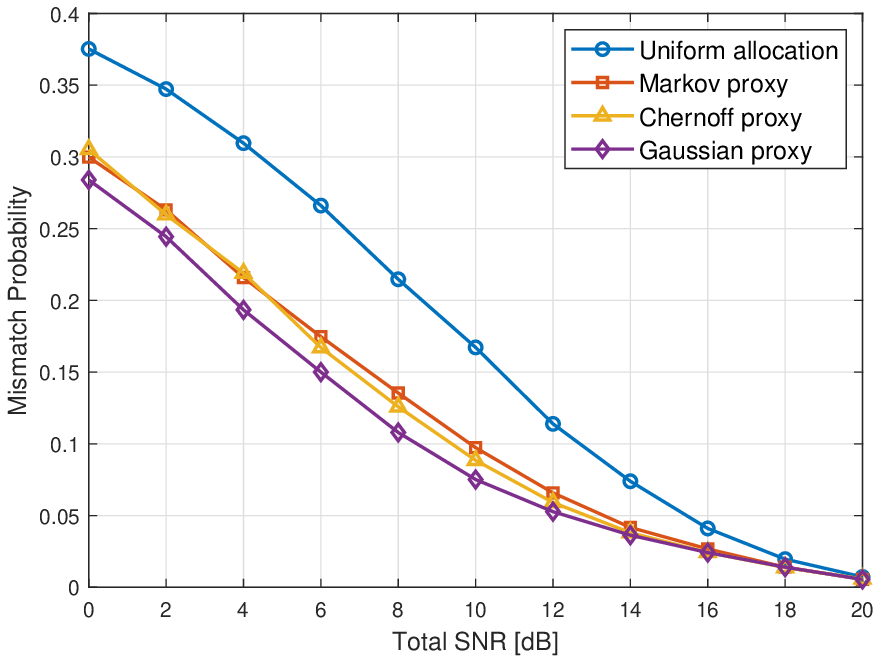}
		\label{fig:T20_pm}}
	\hfil
	\subfloat[]{\includegraphics[width=0.4\textwidth]{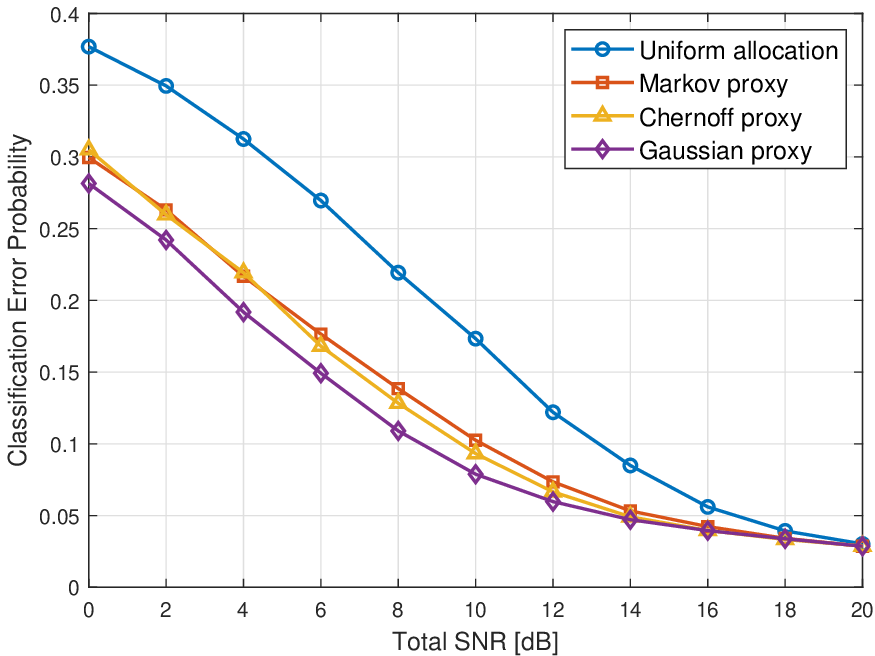}
		\vspace{-4mm}
		\label{fig:T20_pe}}
	\caption{Evaluation of optimized SNRs ($T = 20$): (a) Mismatch probability and (b) classification error probability.}
	\label{fig:T20}
	\end{figure}	
		
	We validate the tools and analytic results of Sections~\ref{sec:proxy} and \ref{sec:resource} with the UCI breast cancer dataset~\cite{Asuncion2007uci}. We compare mismatch probabilities and classification error probabilities of uniform resource allocation and optimized resource allocations for Markov proxy, Chernoff proxy, and Gaussian proxy. The noise-free AdaBoost was trained by Algorithm~\ref{algo:adaboost} with decision stumps as base classifiers. Based on the training output $\boldsymbol{\alpha}$, we compute $\boldsymbol{\beta}$ as shown in Table~\ref{tab:beta} and solve the corresponding optimization problems by \eqref{eq:opt_communication}. 	

	Fig.~\ref{fig:T10} and Fig.~\ref{fig:T20} evaluate the mismatch probabilities and the classification error probabilities of the test set for $T = 10$ and $T = 20$, respectively. We observe that nonuniform communication power allocations can lower the mismatch probability as well as the classification error probability. Among the three nonuniform power allocations (Markov proxy, Chernoff proxy, and Gaussian proxy), the power allocation based on Gaussian proxy achieves the best performance. We emphasize that Fig.~\ref{fig:T10} and Fig.~\ref{fig:T20} plot the actual mismatch and classification error probabilities over the test set optimized with different proxies, and \emph{not} the values of the proxies themselves. Note that the horizontal axis corresponds to the total SNR budget $\frac{\mathcal{C}}{\sigma^2}$. 
	
	For $T = 10$, the Gaussian proxy allocation improves the SNR over the uniform allocation by \SI{3.5}{dB} at $P_e = 0.1$. For $T=20$, the SNR gain is \SI{4.2}{dB} at $P_e = 0.1$. For higher SNR, the mismatch probabilities converge to zero. 
	
	The optimization results by Chernoff proxy are close to the results by Markov proxy in the low-SNR regime. As the SNR increases, the optimization results by Chernoff proxy get close to the results by Gaussian proxy. It can be explained by the Taylor approximation of $\beta = e^{s\alpha} - 1$ in the Chernoff proxy (in Table~\ref{tab:beta}) as follows: 
	\begin{equation}
	e^{s \alpha} - 1 \simeq s \alpha + (s \alpha)^2. 
	\end{equation}
	We observe that the optimal $s^*$ in Algorithm~\ref{algo:chernoff_bound} is small for low SNR and increases for higher SNR. Hence, $\beta$ is approximated to $s \alpha$ in the low SNR region because of $s \alpha \ll 1$. As the SNR increases, $(s \alpha)^2$ is a better approximation to $\beta$. Then, the corresponding coefficients are $\boldsymbol{\beta} = s^2( \alpha_1^2, \ldots, \alpha_T^2)$, which is equivalent to the Gaussian proxy optimization.

	\begin{figure}[t]
		\centering
		\subfloat[]{\includegraphics[width=0.4\textwidth]{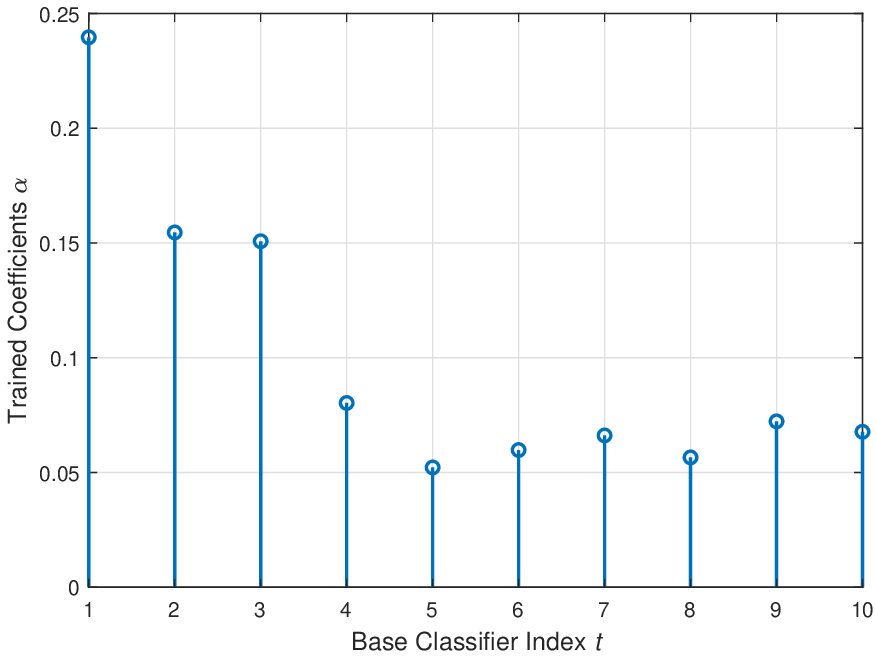}
			\label{fig:alpha_T10}}
		\hfil
		\subfloat[]{\includegraphics[width=0.4\textwidth]{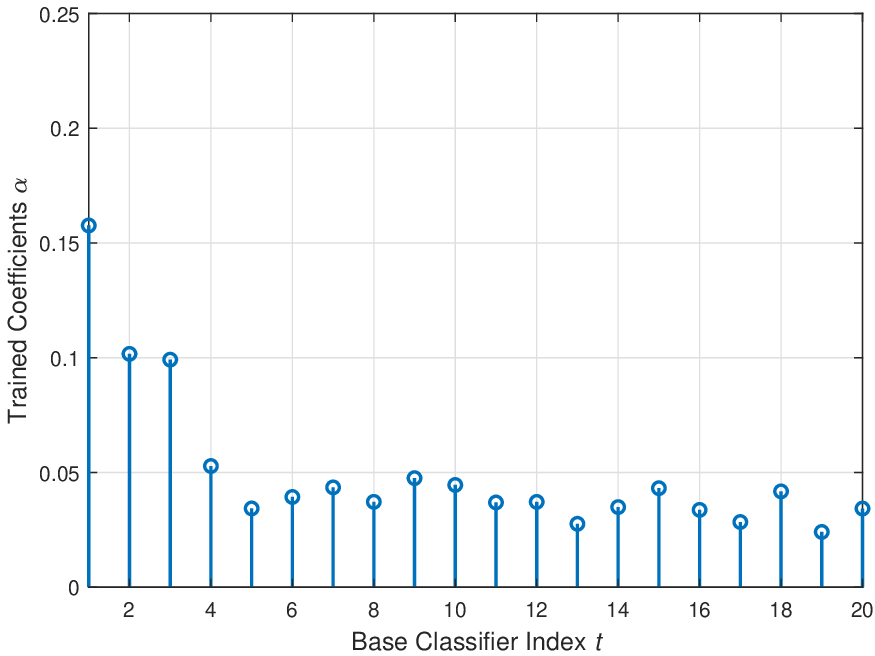}
			\vspace{-4mm}
			\label{fig:alpha_T20}}
		\caption{Trained coefficients $\boldsymbol{\alpha}$: (a) $T = 10$ and (b) $T = 20$.}
		\label{fig:alpha}
	\end{figure}
	
	Fig.~\ref{fig:alpha} shows the trained coefficients $\boldsymbol{\alpha}$. We observe that the values of $\boldsymbol{\alpha}$ are nonuniform. Hence, we can improve the mismatch probability and the classification error probability by allocating the optimized-nonuniform transmit power. We expect that the trained AdaBoost models with lower geometric means of $\boldsymbol{\beta}$ are more robust to the noise as discussed in Corollary~\ref{thm:nonuniform}.  
	
	\begin{figure}[t]
	\centering
	\subfloat[]{\includegraphics[width=0.4\textwidth]{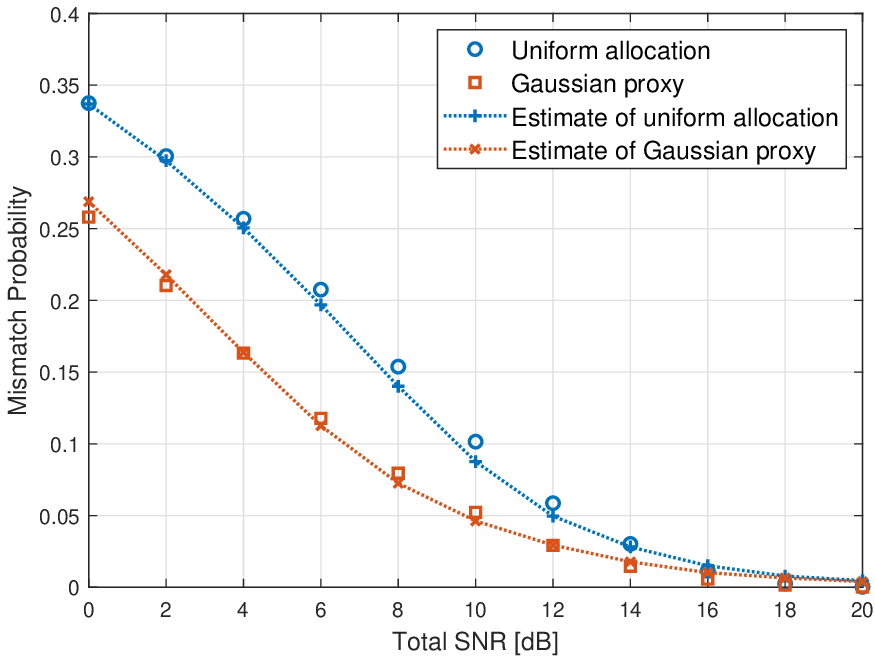}
		\label{fig:est_T10}}
	\hfil
	\subfloat[]{\includegraphics[width=0.4\textwidth]{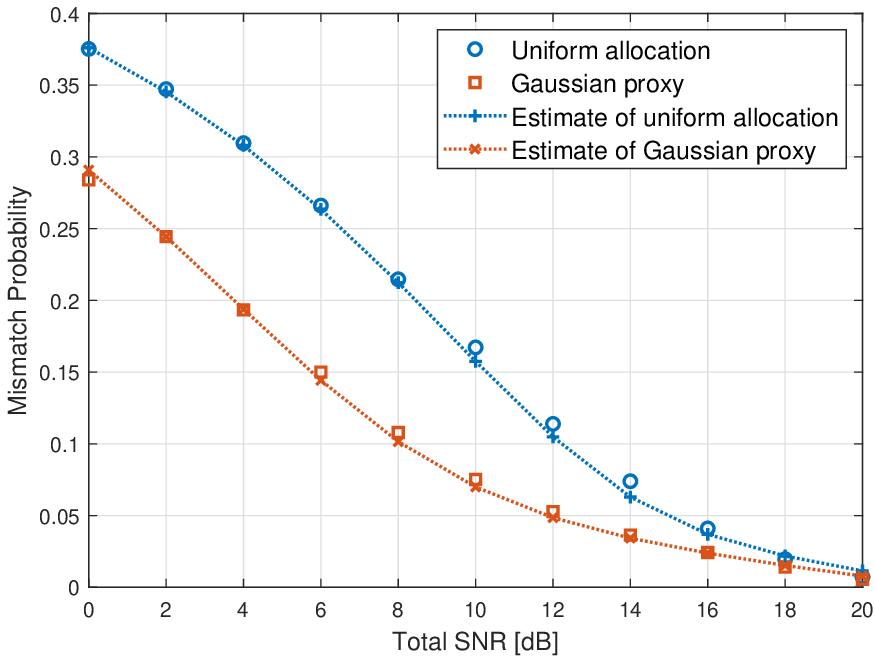}
		\vspace{-4mm}
		\label{fig:est_T20}}
	\caption{Mismatch probabilities and their estimates by \eqref{eq:est_gauss}: (a) $T = 10$ and (b) $T = 20$.}
	\label{fig:est}
	\end{figure}

	Fig.~\ref{fig:est} shows the mismatch probabilities and their estimates for the uniform power allocation and the optimized power allocation using the Gaussian proxy, respectively. The estimates of mismatch probability are calculated by \eqref{eq:est_gauss}. We observe that the estimates match the mismatch probabilities well, which justifies using the Gaussian approximation in Section~\ref{sec:proxy_gaussian}. On the other hand, the other proxies (Markov and Chernoff) have the advantage that their values are proven upper bounds on the mismatch probability (while the Gaussian proxy in general is not a bound).

	\section{Conclusion}\label{sec:conclusion}
	
	In this paper, we propose a principled approach to optimize resource allocation for boosting classifiers. We defined three proxies and the corresponding importance metrics for base classifiers based on Markov inequality, Chernoff bound, and Gaussian approximation. By exploiting the positivity of the importance metrics, we formulated convex resource-allocation problems to minimize the impact of noise. We showed that the proposed approach can effectively improve the classification accuracy for the additive Gaussian noise model. Also, we found that the non-uniform coefficients in boosting offer an advantage over uniform ones (e.g., in bagging) for this noise model. Future work includes different resource/noise models and extension to multi-class noisy inference of boosting classifiers. We believe that the proposed approach can be applied to other settings such as channel-aware distributed detection. 
		
	\appendices
	
	\section{Proof of Theorem~\ref{thm:markov_bound}}\label{pf:markov_bound}
	
	The mismatch probability $P_m(\vect{x}_n)$ is given by
	\begin{align}
	&P_m (\vect{x}_n) \nonumber \\
	&= \Pr\left(  f(\vect{x}_n) > 0 \right) \Pr\left( \widehat{f}(\vect{x}_n ) < 0 \mid f(\vect{x}_n) > 0 \right)  \nonumber \\
	&+ \Pr\left(  f(\vect{x}_n) < 0 \right) \Pr\left( \widehat{f}(\vect{x}_n ) > 0 \mid f(\vect{x}_n) < 0 \right).  
	\end{align}
	
%
	
		

Suppose that $g(\vect{x}_n) = \sum_{t=1}^{T}{\alpha_t f_t(\vect{x}_n)} > 0$. By \eqref{eq:gamma}, 
\begin{equation} \label{eq:gamma_positive}
\sum_{t \in \mathcal{T}^+_n}{\alpha_t} - \sum_{t \in \mathcal{T}^-_n}{\alpha_t} = \gamma_n. 
\end{equation}
where $\mathcal{T}^+_n = \{t \mid f_t(\vect{x}_n) = 1 \}$ and $\mathcal{T}^-_n = \{t \mid f_t(\vect{x}_n) = -1 \}$ $\mathcal{T}^-_n$, respectively. If $f(\vect{x}_n) < 0$, then 
\begin{equation} \label{eq:gamma_negative}
\sum_{t \in \mathcal{T}^+_n}{\alpha_t} - \sum_{t \in \mathcal{T}^-_n}{\alpha_t} = - \gamma_n. 
\end{equation}


	By \eqref{eq:gamma_positive} and \eqref{eq:noisy_f}, $\widehat{f}(\vect{x}_n) < 0$ for given $f(\vect{x}_n) = \gamma_n > 0$ is equivalent to
	\begin{equation} \label{eq:mismatch_pf}
	\sum_{t\in \mathcal{T}^+_n}{\alpha_t \delta_{t,n}} - \sum_{t\in \mathcal{T}^-_n}{\alpha_t \delta_{t,n}} > \frac{\gamma_n}{2}. 
	\end{equation}
	Similarly, $\widehat{f}(\vect{x}_n) > 0$ for given $f(\vect{x}_n) = -\gamma_n < 0$ is equivalent to 
	\begin{equation} \label{eq:mismatch_pf_negative}
	\sum_{t\in \mathcal{T}^+_n}{\alpha_t \delta_{t,n}} - \sum_{t\in \mathcal{T}^-_n}{\alpha_t \delta_{t,n}} < - \frac{\gamma_n}{2}. 	
	\end{equation} 
		
	Hence, the mismatch probability is upper bounded by
	\begin{align}
	P_m (\vect{x}_n) & \le \max\left\{\Pr(f(\vect{x}_n) > 0), \: \Pr(f(\vect{x}_n) < 0)\right\} 
	\nonumber \\
	&\times \Pr\left( \left| \sum_{t\in \mathcal{T}^+_n}{\alpha_t \delta_{t,n}} - \sum_{t\in \mathcal{T}^-_n}{\alpha_t \delta_{t,n}} \right| > \frac{\gamma_n}{2} \right) \label{eq:mismatch_pf_markov_general} \\
	& \le 2 \cdot \frac{\mathbb{E}\left[ \left| \sum_{t\in \mathcal{T}^+_n}{\alpha_t \delta_{t,n}} - \sum_{t\in \mathcal{T}^-_n}{\alpha_t \delta_{t,n}} \right| \right] }{\gamma_n} \label{eq:markov_bound_pf_markov} \\
	& \le 2 \cdot \frac{ \sum_{t=1}^{T}{\alpha_t p_t}}{\gamma_n} \label{eq:markov_bound_pf_1}.
	\end{align}
	where \eqref{eq:markov_bound_pf_markov} follows from the Markov's inequality and \eqref{eq:markov_bound_pf_1} follows from $\mathbb{E}[\delta_{t,n}] = p_t$. 
	
	If $\Pr(f(\vect{x}_n) > 0) = \Pr(f(\vect{x}_n) < 0) = \frac{1}{2}$, then \eqref{eq:markov_bound_pf_1} will be
	\begin{equation} \label{eq:mistmatch_pf_markov_ut}	
	P_m(\vect{x}_n) \le \frac{ \sum_{t=1}^{T}{\alpha_t p_t}}{\gamma_n}. 
	\end{equation}

	\section{Proof of Theorem~\ref{thm:chernoff_bound}}\label{pf:chernoff_bound}	
	
	
	Suppose that $f(\vect{x}_n) > 0$. Then, the mismatch probability $p(\vect{x}_n)$ is given by
	\begin{equation}
	P_m (\vect{x}_n) = \Pr\left( \sum_{t\in \mathcal{T}^+_n}{\alpha_t \delta_{t,n}} - \sum_{t\in \mathcal{T}^-_n}{\alpha_t \delta_{t,n}} > \frac{\gamma_n}{2} \right).  
	\end{equation}
	For any $s > 0$, 
	\begin{align}
	&P_m (\vect{x}_n) \nonumber \\
	&= \Pr\left( e^{s \left(\sum_{t\in \mathcal{T}^+_n}{\alpha_t \delta_{t,n}} - \sum_{t\in \mathcal{T}^-_n}{\alpha_t \delta_{t,n}}\right) } > 
	e^{s\cdot\frac{\gamma_n}{2}} \right) \\
	&\le \frac{\mathbb{E}\left[\exp\left(s \left(\sum_{t\in \mathcal{T}^+_n}{\alpha_t \delta_{t,n}} - \sum_{t\in \mathcal{T}^-_n}{\alpha_t \delta_{t,n}}\right)\right)\right]} {e^{s\cdot\frac{\gamma_n}{2}}} \label{eq:chernoff_bound_pf_markov}
	\end{align}
	where \eqref{eq:chernoff_bound_pf_markov} comes from the Markov's inequality. 
	
	Note that $\delta_{t,n}$ for $t \in \{1,\ldots,T\}$ are independent because of independent $z_t$. Hence, 
	\begin{align}
	&\mathbb{E}\left[\exp\left(s \left\{\sum_{t\in \mathcal{T}^+_n}{\alpha_t \delta_{t,n}} - \sum_{t\in \mathcal{T}^-_n}{\alpha_t \delta_{t,n}}\right\}\right)\right] \\
	&= \prod_{t\in \mathcal{T}^+_n}{\mathbb{E}\left[ \exp\left(s \alpha_t \delta_{t,n}\right) \right]} \cdot \prod_{t\in \mathcal{T}^-_n}{\mathbb{E}\left[ \exp\left(- s \alpha_t \delta_{t,n}\right) \right]}. \label{eq:chernoff_bound_pf_prod}
	\end{align}  
	In addition,
	\begin{align}
	\mathbb{E}\left[ \exp\left( s \alpha_t \delta_{t,n} \right) \right] &= 1 + p_t \left\{ \exp(s\alpha_t) - 1 \right\} \\
	& \le \exp\left( p_t \left( e^{s \alpha_t} - 1 \right) \right) \label{eq:chernoff_bound_pf_exp}
	\end{align}
	where \eqref{eq:chernoff_bound_pf_exp} follows from $1 + u \le \exp(u)$ and $u = p_t \left( e^{s \alpha_t} - 1 \right)$. Similarly, $\mathbb{E}\left[ \exp\left( - s \alpha_t \delta_{t,n} \right) \right] \le \exp\left( - p_t \left(1 - e^{ - s \alpha_t} \right) \right)$. 
	
	By \eqref{eq:chernoff_bound_pf_markov}, \eqref{eq:chernoff_bound_pf_prod}, and \eqref{eq:chernoff_bound_pf_exp}, 
	\begin{align}
	P_m (\vect{x}_n)
	&\le \exp\left( - s\cdot \frac{\gamma_n}{2} + \sum_{t\in \mathcal{T}^+_n}{  p_t \left( e^{s \alpha_t} - 1 \right) }  \right. \nonumber \\ 
	& \quad \left. - \sum_{t\in \mathcal{T}^-_n}{  p_t \left(1 -  e^{- s \alpha_t} \right) } \right) \\
	&\le \exp\left( - s\cdot \frac{\gamma_n}{2} + \sum_{t=1}^{T}{  p_t \left( e^{s \alpha_t} - 1 \right) } \right). \label{eq:chernoff_bound_pf_total0} 	
	\end{align}
	Similarly, we can obtain the same upper bound for a case of $\sum_{t=1}^{T}{\alpha_t f_t(\vect{x}_n)} = -\gamma_n < 0$. 
	
	Hence, the upper bound on the mismatch probability is given by
	\begin{align}
	P_m &= \mathbb{E}\left[ P_m(\vect{x}_n) \right] \nonumber \\
	&\le \mathbb{E}\left[ \exp\left(-s \cdot \frac{\gamma_n}{2}\right) \right] \cdot \exp\left(\sum_{t=1}^{T}{  p_t \left( e^{s \alpha_t} - 1 \right) } \right) \\
	&= \mathbb{E}\left[ \exp\left(-s \cdot \frac{\gamma_n}{2}\right) \right] \cdot \exp\left( \widehat{p}_C \right)
	\end{align}
	where a larger $s$ reduces $\mathbb{E}\left[ \exp\left(-s \cdot \frac{\gamma_n}{2}\right) \right]$ while it increases $\exp\left( \widehat{p}_C \right)$.

\section{Proof of Theorem~\ref{thm:optimal_s}}\label{pf:optimal_s}	
	For given $\vect{p} = (p_1, \ldots, p_T)$ and ${\boldsymbol \gamma}=(\gamma_1, \ldots, \gamma_N)$, the upper bound on $P_m$ is given by
	\begin{align}
	h(s) & = \left(\frac{1}{N} \sum_{n=1}^{N}{e^{-s \cdot \frac{\gamma_n}{2}}} \right) \cdot \exp\left( \sum_{t=1}^{T}{ \left(e^{s \alpha_t} - 1\right)p_t} \right) \nonumber \\
	&= \frac{1}{N} \cdot h_1(s) \cdot h_2(s)
	\end{align}
	where 
	\begin{align}
		h_1(s) &= \sum_{n=1}^{N}{e^{-s \cdot \frac{\gamma_n}{2}}}, \\
		h_2(s) &= \exp\left( \sum_{t=1}^{T}{ \left(e^{s \alpha_t} - 1\right)p_t} \right). 
	\end{align}
	
	We show that $h''(s) \ge 0$ to check the convexity of $h(s)$. First, we obtain
	\begin{align}
	h'(s) &= h_1'(s) h_2(s) + h_1(s) h_2'(s) \\
	&= \sum_{n=1}^{N}{\left\{  \left(\sum_{t=1}^{T}{p_t \alpha_t e^{s \alpha_t}} -\frac{\gamma_n}{2}\right) e^{-s \cdot \frac{\gamma_n}{2}}\right\}} \cdot  h_2(s)
	\end{align}
	where $h_2(s) > 0$ for $s>0$ and $\alpha_t > 0$. Hence, $h'(s) = 0$ can be achieved by $s$ satisfying 
	\begin{equation} \label{eq:pf_optimal_s}
	\sum_{n=1}^{N}{\left\{  \left(\sum_{t=1}^{T}{p_t \alpha_t e^{s \alpha_t}} -\frac{\gamma_n}{2}\right) e^{-s \cdot \frac{\gamma_n}{2}} \right\}} = 0
	\end{equation}
	which is equivalent to \eqref{eq:algo_s} in Algorithm~\ref{algo:chernoff_bound}. 
	
	The second derivative of $h(s)$ is given by
	\begin{align}
	h''(s) &= h_2(s) \cdot \sum_{n=1}^{N}{ e^{-s \cdot \frac{\gamma_n}{2}} \cdot \frac{h_3(s)}{4}}  
	\end{align}
	where $h_3(s)$ is 
	\begin{align}
	h_3(s) &= \gamma_n^2 - 4 \gamma_n \sum_{t=1}^{T}{p_t \alpha_t e^{s\alpha_t}} \nonumber \\
	& + 4 \left\{\sum_{t=1}^{T}{p_t \alpha_t^2 e^{s\alpha_t}} + \left(\sum_{t=1}^{T}{p_t \alpha_t e^{s\alpha_t}} \right)^2 \right\} \\ 
	&= \left(\gamma_n - 2 \gamma_n \sum_{t=1}^{T}{p_t \alpha_t e^{s\alpha_t}} \right)^2 + 4\sum_{t=1}^{T}{p_t \alpha_t^2 e^{s\alpha_t}}. 
	\end{align}
	Note that $h_2(s) >0$ and $h_3(s) > 0$ for $s> 0$, $\alpha_t \ge 0$ and $p_t \ge 0$ for any $t$. Hence, $h''(s)>0$ and the optimal $s$ should satisfy \eqref{eq:pf_optimal_s}. 

\section{Proof of Theorem~\ref{thm:gaussian}}\label{pf:gaussian}

By~\eqref{eq:signal_noise}, we showed that $\widehat{f}(\vect{x}_n) = \pm \gamma_n + v_n$. The classification noise term $v_n$ is given by
\begin{equation}
v_n = - 2 \left(\sum_{t\in \mathcal{T}^+_n}{\alpha_t \delta_{t,n}} - \sum_{t\in \mathcal{T}^-_n}{\alpha_t \delta_{t,n}} \right). 
\end{equation} 

Since we assume that $z_t$s of \eqref{eq:noisy_base} are independent, $\delta_{t,n}$s are also independent for a given $\vect{x}_n$. By the central limit theorem, $\sum_{t\in \mathcal{T}^+_n}{\alpha_t p_t}$ and $\sum_{t\in \mathcal{T}^-_n}{\alpha_t p_t}$ can be approximated as Gaussian distributions, respectively. Hence, $v_n$ can be modeled as Gaussian distribution as well. 

The mean of $v_n$ is readily derived by using $\mathbb{E}[\delta_{t,n}] = p_t$. The variance of $v_n$ is given by
\begin{align}
\sigma_v^2 &= 4 \sum_{t=1}^{T}{\alpha_t^2 \cdot \text{Var}[\delta_{t,n}]} \\
& = 4 \sum_{t=1}^{T}{\alpha_t^2 \cdot p_t (1 - p_t)} \label{eq:v_var_pf}
\end{align}
where $\text{Var}[\delta_{t,n}] =\mathbb{E} [\delta_{t,n}^2] - \mu_v^2 =   p_t (1-p_t)$. 
	
%
%
%
%
%

\section{Proof of Theorem~\ref{thm:opt_communication_sol}}\label{pf:opt_communication_sol}

The Lagrangian $L(\vect{r}, \nu, \boldsymbol{\lambda})$ of \eqref{eq:opt_communication} is given by
\begin{align}
& L(\vect{r}, \nu, \boldsymbol{\lambda}) \nonumber \\
&= \sum_{t=1}^{T}{\beta_t Q \left(\frac{r_t}{\sigma_t}\right)} + \nu \left(\sum_{t=1}^{T}{r_t^2} - \mathcal{C}\right) - \sum_{t=1}^{T}{\lambda_t r_t}. 
\end{align}
The corresponding KKT conditions are as follows:
\begin{align}
\sum_{t=1}^{T}{r_t^2} &\le \mathcal{C}, \quad \nu \ge 0, \quad
\nu \cdot \left(\sum_{t=1}^{T}{r_t^2} - \mathcal{C}\right) = 0, \label{eq:cr1_KKT_1} \\
r_t &\ge 0, \quad \lambda_t \ge 0, \quad \lambda_t r_t  = 0 \label{eq:cr1_KKT_2} \\
\frac{\partial L}{\partial r_t} &= - \frac{\beta_t}{\sqrt{2\pi} \sigma_t} \exp\left(-\frac{r_t^2}{2 \sigma_t^2}\right) + 2\nu r_t - \lambda_t= 0 \label{eq:cr1_KKT_3}
\end{align}
for $t \in \{1,\ldots, T\}$. 

From \eqref{eq:cr1_KKT_3}, $\lambda_t$ is
\begin{equation} \label{eq:cr1_KKT_lambda}
\lambda_t = 2\nu r_t - \frac{\beta_t}{\sqrt{2\pi}\sigma_t} \exp\left(-\frac{r_t^2}{2 \sigma_t^2}\right). 
\end{equation}
If $\nu = 0$. Then, $\lambda_t < 0$, which violates \eqref{eq:cr1_KKT_2} because of $\beta_t > 0$. Hence, we claim that $\nu \ne 0$, which results in $\sum_{t=1}^{T}{r_t^2} = \mathcal{C}$. 

If $r_t = 0$, then $\lambda_t =  - \frac{\beta_t}{\sqrt{2\pi}\sigma_t}$, which violates \eqref{eq:cr1_KKT_2}. Hence, we claim that $r_t > 0$ and $\lambda_t = 0$, i.e., 
\begin{equation}\label{eq:cr1_KKT_slack_1}
r_t = \frac{\beta_t}{2 \sqrt{2\pi}\sigma_t \nu} \exp\left(-\frac{r_t^2}{2 \sigma_t^2} \right), 
\end{equation}
which is equivalent to
\begin{equation}
\frac{r_t^2}{\sigma_t^2} \exp\left(\frac{r_t^2}{\sigma_t^2}\right) = \frac{\beta_t^2}{8\pi \sigma_t^4 \nu^2}. 
\end{equation}
By setting $x = \frac{r_t^2}{\sigma_t^2}$, we obtain $x \exp(x) = \frac{\beta_t^2}{8\pi \sigma_t^4 \nu^2}$. Hence, $x = \mathsf{SNR_t} = W\left(\frac{\beta_t^2}{8\pi \sigma_t^4 \nu^2} \right)$. 

\section{Proof of Theorem~\ref{thm:nonuniform}}\label{pf:nonuniform}

By replacing $Q\left( x \right)$ with its Chernoff bound $\frac{1}{2}\exp\left(-\frac{x^2}{2}\right)$, the optimization problem \eqref{eq:opt_communication} will be modified to 
\begin{equation}
\begin{aligned} \label{eq:approx_opt}
& \underset{\vect{x}}{\text{minimize}}
& & \sum_{t=1}^{T}{\frac{\beta_t}{2} \exp\left(-\frac{x_t^2}{2}\right)} \\
&{\text{subject~to}} & & \sum_{t=1}^{T}{x_t^2} \le \frac{\mathcal{C}}{\sigma^2} \\
&                    & & x_t \ge 0 \quad t=1,\ldots,T
\end{aligned}
\end{equation}
where $x_t = \frac{r_t}{\sigma_t}$. The corresponding Lagrangian is given by
\begin{align}
L(\vect{x}, \nu, \boldsymbol{\lambda}) &= \sum_{t=1}^{T}{\frac{\beta_t}{2} \exp\left(-\frac{x_t^2}{2}\right)} \nonumber \\
& + \nu \left(\sum_{t=1}^{T}{x_t^2} - \mathcal{C}'\right) - \sum_{t=1}^{T}{\lambda_t x_t}. 
\end{align}

From the KKT conditions, we can obtain the following conditions of the optimal $x^*$ (as in Appendix~\ref{pf:opt_communication_sol}):
\begin{align}
\sum_{t=1}^{T}{(x_t^*)^2} &= \frac{\mathcal{C}}{\sigma^2} \label{eq:approx_con0} \\
\frac{\beta_t}{2} \exp\left(-\frac{(x_t^*)^2}{2}\right) &= 2 \nu  \label{eq:approx_con1}.   
\end{align}
By \eqref{eq:approx_opt} and \eqref{eq:approx_con1}, the optimized proxy is given by
\begin{align}
\sum_{t=1}^{T}{\frac{\beta_t}{2} \exp\left(-\frac{(x_t^*)^2}{2}\right)} = 2T\nu. 
\end{align}

Note that \eqref{eq:approx_con1} is equivalent to
\begin{equation} \label{eq:approx_con2}
(x_t^*)^2 = \mathsf{SNR}_t^* = 2 \log\frac{\beta_t}{4 \nu}. 
\end{equation}
By \eqref{eq:approx_con0} and \eqref{eq:approx_con2}, 
\begin{align}
\log{\prod_{t=1}^{T}{\frac{\beta_t}{4\nu}}} = \frac{\mathcal{C}}{2\sigma^2}, 
\end{align}
which leads to $\nu = \frac{1}{4}\exp\left(-\frac{\mathcal{C}}{2T\sigma^2}\right)\left(\prod_{t=1}^{T}{\beta_t}\right)^{\frac{1}{T}}$. Then, the optimized proxy is given by
\begin{equation}
2T\nu = \frac{T}{2}\exp\left(-\frac{\mathcal{C}}{2T\sigma^2}\right)\left(\prod_{t=1}^{T}{\beta_t}\right)^{\frac{1}{T}}. 
\end{equation}
		
	

	%
	
	
	
	\bibliographystyle{IEEEtran}
	\bibliography{abrv,mybib}

\begin{thebibliography}{10}
\providecommand{\url}[1]{#1}
\csname url@samestyle\endcsname
\providecommand{\newblock}{\relax}
\providecommand{\bibinfo}[2]{#2}
\providecommand{\BIBentrySTDinterwordspacing}{\spaceskip=0pt\relax}
\providecommand{\BIBentryALTinterwordstretchfactor}{4}
\providecommand{\BIBentryALTinterwordspacing}{\spaceskip=\fontdimen2\font plus
\BIBentryALTinterwordstretchfactor\fontdimen3\font minus
  \fontdimen4\font\relax}
\providecommand{\BIBforeignlanguage}[2]{{%
\expandafter\ifx\csname l@#1\endcsname\relax
\typeout{** WARNING: IEEEtran.bst: No hyphenation pattern has been}%
\typeout{** loaded for the language `#1'. Using the pattern for}%
\typeout{** the default language instead.}%
\else
\language=\csname l@#1\endcsname
\fi
#2}}
\providecommand{\BIBdecl}{\relax}
\BIBdecl

\bibitem{Dietterich2000ensemble}
T.~G. Dietterich, ``{Ensemble methods in machine learning},'' in \emph{Proc.
  Int. Workshop Multiple Classifier Syst.}, Dec. 2000, pp. 1--15.

\bibitem{Freund1997decision}
Y.~Freund and R.~E. Schapire, ``{A decision-theoretic generalization of on-line
  learning and an application to boosting},'' \emph{J. Comput. Syst. Sci.},
  vol.~55, no.~1, pp. 119--139, Dec. 1997.

\bibitem{Bishop2006pattern}
C.~M. Bishop, \emph{{Pattern Recognition and Machine Learning}}.\hskip 1em plus
  0.5em minus 0.4em\relax New York, NY, USA: Springer, 2006.

\bibitem{Freund1996experiments}
Y.~Freund and R.~E. Schapire, ``{Experiments with a new boosting algorithm},''
  in \emph{Proc. Int. Conf. Mach. Learn. (ICML)}, Jul. 1996, pp. 148--156.

\bibitem{Viola2001rapid}
P.~Viola and M.~Jones, ``{Rapid object detection using a boosted cascade of
  simple features},'' in \emph{Proc. IEEE Conf. Comput. Vis. Pattern
  Recognition (CVPR)}, Dec. 2001, pp. I--511--I--518.

\bibitem{Friedman2001greedy}
J.~H. Friedman, ``{Greedy function approximation: A gradient boosting
  machine},'' \emph{Ann. Stat.}, vol.~29, no.~5, pp. 1189--1232, Oct. 2001.

\bibitem{Breiman1996bagging}
L.~Breiman, ``{Bagging predictors},'' \emph{Mach. Learn.}, vol.~24, no.~2, pp.
  123--140, Aug. 1996.

\bibitem{Shanbhag2019shannon}
N.~R. Shanbhag, N.~Verma, Y.~Kim, A.~D. Patil, and L.~R. Varshney,
  ``{Shannon-inspired statistical computing for the nanoscale era},''
  \emph{Proc. {IEEE}}, vol. 107, no.~1, pp. 90--107, Jan. 2019.

\bibitem{Poor1977applications}
H.~V. Poor and J.~B. Thomas, ``{Applications of Ali-Silvey distance measures in
  the design of generalized quantizers for binary decision systems},''
  \emph{{IEEE} Trans. Commun.}, vol.~25, no.~9, pp. 893--900, Sep. 1977.

\bibitem{Sakr2017analytical}
C.~Sakr, Y.~Kim, and N.~Shanbhag, ``{Analytical guarantees on numerical
  precision of deep neural networks},'' in \emph{Proc. Int. Conf. Mach. Learn.
  (ICML)}, Aug. 2017, pp. 3007--3016.

\bibitem{Sakr2017minimum}
C.~Sakr, A.~Patil, S.~Zhang, Y.~Kim, and N.~Shanbhag, ``{Minimum precision
  requirements for the SVM-SGD learning algorithm},'' in \emph{Proc. IEEE Int.
  Conf. Acoust., Speech, Signal Process. (ICASSP)}, Mar. 2017, pp. 1138--1142.

\bibitem{Neumann1956probabilistic}
J.~{von Neumann}, ``{Probabilistic logics and the synthesis of reliable
  organisms from unreliable components},'' \emph{Automata Studies}, vol.~34,
  pp. 43--98, 1956.

\bibitem{Donmez2016cost}
M.~A. Donmez, M.~Raginsky, A.~C. Singer, and L.~R. Varshney,
  ``{Cost-performance tradeoffs in unreliable computation architectures},'' in
  \emph{Proc. Asilomar Conf. Signals, Syst. Comput.}, Nov. 2016, pp. 215--219.

\bibitem{Angluin1988learning}
D.~Angluin and P.~Laird, ``{Learning from noisy examples},'' \emph{Mach.
  Learn.}, vol.~2, no.~4, pp. 343--370, Apr. 1988.

\bibitem{Frenay2014classification}
B.~Frenay and M.~Verleysen, ``{Classification in the presence of label noise: A
  survey},'' \emph{{IEEE} Trans. Neural Netw.}, vol.~25, no.~5, pp. 845--869,
  May 2014.

\bibitem{Dietterich2000experimental}
T.~G. Dietterich, ``{An experimental comparison of three methods for
  constructing ensembles of decision trees: Bagging, boosting, and
  randomization},'' \emph{Mach. Learn.}, vol.~40, no.~2, pp. 139--157, Aug.
  2000.

\bibitem{Domingo2000madaboost}
C.~Domingo and O.~Watanabe, ``{MadaBoost: A modification of adaBoost},'' in
  \emph{Proc. Annu. Conf. Comput. Learn. Theory (COLT)}, Jun.-Jul. 2000, pp.
  180--189.

\bibitem{Long2010random}
P.~M. Long and R.~A. Servedio, ``{Random classification noise defeats all
  convex potential boosters},'' \emph{Mach. Learn.}, vol.~78, no.~3, pp.
  287--304, Mar. 2010.

\bibitem{Mohri2018foundation}
M.~Mohri, A.~Rostamizadeh, and A.~Talwalkar, \emph{{Foundations of Machine
  Learning}}, 2nd~ed.\hskip 1em plus 0.5em minus 0.4em\relax Cambridge, MA,
  USA: MIT Press, 2018.

\bibitem{Wang2015eacb}
Z.~Wang, R.~E. Schapire, and N.~Verma, ``{Error adaptive classifier boosting
  (EACB): Leveraging data-driven training towards hardware resilience for
  signal inference},'' \emph{{IEEE} Trans. Circuits Syst. {I}}, vol.~62, no.~4,
  pp. 1136--1145, Apr. 2015.

\bibitem{Wang2015overcoming}
Z.~Wang, K.~H. Lee, and N.~Verma, ``{Overcoming computational errors in sensing
  platforms through embedded machine-learning kernels},'' \emph{{IEEE} Trans.
  {VLSI} Syst.}, vol.~23, no.~8, pp. 1459--1470, Aug. 2015.

\bibitem{Chair1986optimal}
Z.~Chair and P.~K. Varshney, ``{Optimal data fusion in multiple sensor
  detection systems},'' \emph{{IEEE} Trans. Aerosp. Electron. Syst.}, vol.
  AES-22, no.~1, pp. 98--101, Jan. 1986.

\bibitem{Viswanathan1997distributed}
R.~Viswanathan and P.~K. Varshney, ``{Distributed detection with multiple
  sensors: Part I--fundamentals},'' \emph{Proc. {IEEE}}, vol.~85, no.~1, pp.
  54--63, Jan. 1997.

\bibitem{Li2008adaboost}
X.~Li, L.~Wang, and E.~Sung, ``{AdaBoost with SVM-based component
  classifiers},'' \emph{Eng. Appl. Artif. Intell.}, vol.~21, no.~5, pp.
  785--795, Sep. 2008.

\bibitem{Schwenk1997training}
H.~Schwenk and Y.~Bengio, ``{Training methods for adaptive boosting of neural
  networks},'' in \emph{Proc. Annu. Conf. Neural Inf. Process. Syst. (NIPS)},
  Dec. 1997, pp. 647--650.

\bibitem{Schwenk2000boosting}
------, ``{Boosting neural networks},'' \emph{Neural Comput.}, vol.~12, no.~8,
  pp. 1869--1887, Aug. 2000.

\bibitem{Chen2005optimality}
B.~Chen and P.~K. Willett, ``{On the optimality of the likelihood-ratio test
  for local sensor decision rules in the presence of nonideal channels},''
  \emph{{IEEE} Trans. Inf. Theory}, vol.~51, no.~2, pp. 693--699, Feb. 2005.

\bibitem{Chen2006channel}
B.~Chen, L.~Tong, and P.~K. Varshney, ``{Channel-aware distributed detection in
  wireless sensor networks},'' \emph{{IEEE} Signal Process. Mag.}, vol.~23,
  no.~4, pp. 16--26, Jul. 2006.

\bibitem{Corless1996lambert}
R.~M. Corless, G.~H. Gonnet, D.~E.~G. Hare, D.~J. Jeffrey, and D.~E. Knuth,
  ``{On the Lambert W function},'' \emph{Adv. Comput. Math.}, vol.~5, no.~1,
  pp. 329--359, Dec. 1996.

\bibitem{Asuncion2007uci}
\BIBentryALTinterwordspacing
A.~Asuncion and D.~Newman, ``{UCI machine learning repository},'' 2007.
  [Online]. Available: \url{http://archive.ics.uci.edu/ml}
\BIBentrySTDinterwordspacing

\end{thebibliography}

\end{document}